\documentclass[accepted]{uai2021} 

\usepackage[american]{babel}

\usepackage{natbib} 
\usepackage{mathtools} 
\usepackage{booktabs} 
\usepackage{tikz} 

\usepackage{subcaption}
\usepackage{graphicx}
\usepackage{pgf}
\usepackage{amsthm}
\usepackage{color}

\usepackage{amsmath,amsfonts,bm}

\def\eqref#1{equation~\ref{#1}}

\def\1{\bm{1}}

\DeclareMathAlphabet{\mathsfit}{\encodingdefault}{\sfdefault}{m}{sl}
\SetMathAlphabet{\mathsfit}{bold}{\encodingdefault}{\sfdefault}{bx}{n}

\newcommand{\R}{\mathbb{R}}

\renewcommand{\d}{\mathrm{d}}

\newtheorem{theorem}{Theorem}
\newtheorem{lemma}{Lemma}

\title{Identifying Untrustworthy Predictions in Neural Networks\\by Geometric Gradient Analysis}

\author[1]{\href{mailto:Leo Schwinn <leo.schwinn@fau.de>?Subject=Geometric Gradient Analysis 2021 paper 2021}{Leo~Schwinn}{}}
\author[1]{An~Nguyen}
\author[1]{Ren\'e~Raab}
\author[2]{Leon~Bungert}
\author[2]{Daniel~Tenbrinck}
\author[1]{Dario~Zanca}
\author[2]{Martin~Burger}
\author[1]{Bjoern~Eskofier}
\affil[1]{%
    Department Artificial Intelligence in Biomedical Engineering\\
    Univ. of Erlangen-Nürnberg\\
    Germany
}
\affil[2]{%
    Department Mathematics\\
    Univ. of Erlangen-Nürnberg\\
    Germany
}

\begin{document}
\maketitle

\begin{abstract}

The susceptibility of deep neural networks to untrustworthy predictions, including out-of-distribution (OOD) data and adversarial examples, still prevent their widespread use in safety-critical applications. Most existing methods either require a re-training of a given model to achieve robust identification of adversarial attacks or are limited to out-of-distribution sample detection only. In this work, we propose a geometric gradient analysis (GGA) to improve the identification of untrustworthy predictions without retraining of a given model. GGA analyzes the geometry of the loss landscape of neural networks based on the saliency maps of their respective input. To motivate the proposed approach, we provide theoretical connections between gradients' geometrical properties and local minima of the loss function. Furthermore, we demonstrate that the proposed method outperforms prior approaches in detecting OOD data and adversarial attacks, including state-of-the-art and adaptive attacks.

\end{abstract}

\section{Introduction}

Deep neural networks (DNNs) are known to achieve remarkable results when the distribution of the training and test data are similar. However, this assumption is often violated in real-world scenarios where so-called out-of-distribution (OOD) data may be observed which are not covered by the training set. DNNs have been shown to make high-confidence predictions for OOD data even if it does not contain any semantic information, e.g., randomly generated noise \citep{Hendrycks17}. This behaviour can lead to fatal outcomes in safety-critical applications, for example in autonomous driving, where the algorithm might fail to call for human intervention when it is confronted with OOD data. In addition to the overconfidence of DNNs, it is widely recognized that most DNNs are vulnerable to imperceptible input perturbations called adversarial examples \citep{Goodfellow2015,Madry2018}. These perturbations can lead to incorrect predictions by the neural network and therefore pose an additional security risk. Many approaches have been proposed to make neural networks more robust in terms of adversarial examples \citep{Goodfellow2015,Madry2018,Gowal2020Robustness}. Nevertheless, there is still a wide gap between the accuracy on unperturbed data and adversarial examples. 

An alternative to training robust DNNs is the early detection of attacks \citep{Lee18,Chen2020robust}. Identified attacks can then be forwarded for further human assessment. One line of research investigates geometric properties of neural networks in the input space to explain their classification decisions and detect adversarial attacks. \citet{Fawzi17ClassificationRegions} demonstrate that the decision boundaries of neural networks are mostly flat around the training data and only show considerable curvature in very few directions. \citet{Jetley18Losslandscape} illustrate that these high-curvature directions are mainly responsible for the final classification decision and thus can be exploited by adversarial attacks to induce misclassifications. However, \citet{Fawzi17ClassificationRegions} only focus on detecting small adversarial perturbations and \citet{Jetley18Losslandscape} restrict themselves to the theoretical analysis of the loss landscape.

\begin{figure*}[t]
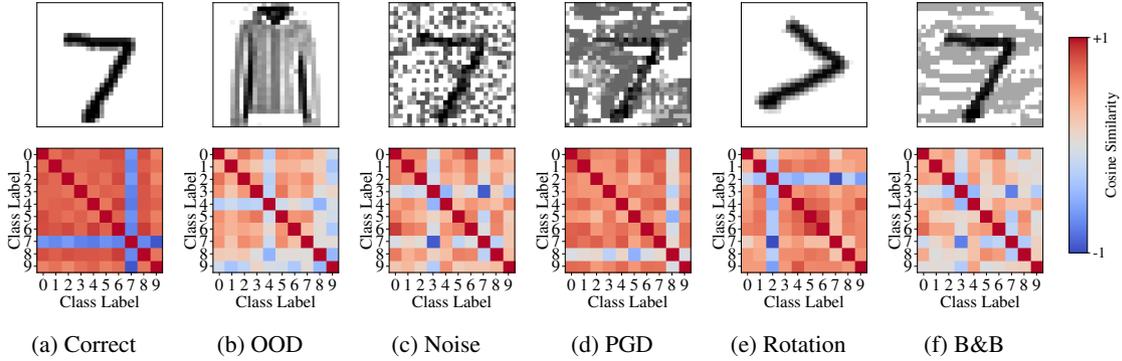

\centering
  \begin{subfigure}{0.13\textwidth}
    \resizebox{\textwidth}{!}{\input{Images/CS_Correct.pgf}}
    \caption{Correct}
    \label{fig:gradient_geometry_a}
  \end{subfigure}
  \begin{subfigure}{0.13\textwidth}
    \resizebox{\textwidth}{!}{\input{Images/CS_OOD.pgf}}
    \caption{OOD}
  \end{subfigure}
  \begin{subfigure}{0.13\textwidth}
    \resizebox{\textwidth}{!}{\input{Images/CS_Noise.pgf}}
    \caption{Noise}
  \end{subfigure}
  \begin{subfigure}{0.13\textwidth}
    \resizebox{\textwidth}{!}{\input{Images/CS_PGD.pgf}}
    \caption{PGD}
  \end{subfigure}
  \begin{subfigure}{0.13\textwidth}
    \resizebox{\textwidth}{!}{\input{Images/CS_Rotation.pgf}}
    \caption{Rotation}
  \end{subfigure}
  \begin{subfigure}{0.13\textwidth}
    \resizebox{\textwidth}{!}{\input{Images/CS_BB.pgf}}
    \caption{B\&B}
  \end{subfigure}
  \begin{subfigure}{0.08\textwidth}
    \resizebox{\textwidth}{!}{\input{Images/colorbar.pgf}}
  \end{subfigure}
 
 \caption{Input samples of a neural network in the top row and the respective cosine similarity matrices in the bottom row (matrices which contain the pairwise cosine similarity between the saliency map w.r.t.\ every class-specific logit). Correct classification results in high average cosine similarity (red) between the saliency maps of non-predicted classes while adversarial attacks and outliers can be detected by saliency maps that are less aligned between non-predicted-classes (blue).}
 \label{fig:gradient_geometry}
\end{figure*}

In this work we focus on the detection of two major problems of DNNs, namely OOD data and adversarial attacks. We propose a novel methodology inspired by the analysis of geometric properties in the input space of neural networks, which we name geometric gradient analysis (GGA). Here, we analyze and interpret the gradient of a neural network with respect to its input, in the following referred to as \emph{saliency map}. More precisely, for a given input sample we inspect the geometric relation among all possible saliency maps, calculated for each output class of the model. This is achieved by a pairwise calculation of the cosine similarity between saliency maps. GGA can be used with any pre-trained differentiable neural network and does not require any re-training of the model. Figure~\ref{fig:gradient_geometry} shows input samples of a neural network in the top row and the respective cosine similarities between the saliency maps of every output class in the bottom row. Figure~\ref{fig:gradient_geometry_a} exemplifies that if an input is correctly classified by the model, the saliency map of the predicted class (i.e. the digit $7$) generally points in a direction which is opposite to the saliency maps of all other classes. This results in low average cosine similarity between these saliency maps in the rows and columns of class label $7$ (blue colored squares). Accordingly, the saliency maps of the other classes mostly align and display a high average cosine similarity (red colored squares). In contrast, in the case of a OOD sample or adversarial attack, the saliency maps of the non-predicted classes point towards different directions and the cosine similarity is considerably lower on average.
The contributions of this paper can be summarized as follows.
First, we theoretically motivate the proposed GGA method by deriving a connection between the local minima of the loss landscape and the geometric behavior of gradients. Subsequently, we quantitatively demonstrate that for common OOD tasks GGA is highly competitive compared to prior methods. Furthermore, we demonstrate that GGA successfully identifies a diverse variety of adversarial attacks and show that the geometric relation between gradients is difficult to compromise with adaptive attacks. 

\section{Related Work}

Our proposed method combines ideas from several areas of neural network research. This includes out-of-distribution detection, robust out-of-distribution detection which combines OOD detection and adversarial attacks, and model saliency. In this section, we briefly review prior work in these research areas.

Previous works have established that softmax-based neural networks tend to make overconfident predictions in the presence of misclassifications, OOD data, and adversarial attacks \citep{Hendrycks17,Liang18,Jiang18Misclassification,Corbiere19Misclassifications}. 
\citet{Hendrycks17} propose a baseline method for detecting OOD data which utilizes the softmax output of a neural network. Depending on a pre-defined threshold based on the softmax score they define samples as either in- or out-of-distribution. \citet{Liang18} further enhance this baseline. They apply temperature scaling to the softmax scores and additionally add small perturbations to the input to increase the difference between in- and out-of-distribution samples. While both approaches have been shown to work on OOD data, they fail in the presence of adversarial attacks \citep{Chen2020robust}. 

\citet{Lee18} evaluate their detection framework both on OOD and adversarial samples. They calculate class-conditional Gaussian distributions from the pre-trained networks and discriminate samples based on the Mahalanobis distance between the distributions. \citet{Chen2020robust} proposed a combined framework for detecting OOD data and adversarial attacks as robust out-of-distribution (ROOD) detection. They extend the threat model to attacks on OOD data which aim to fool the adversarial detector as well as the classification model. They augment the training data of neural networks with both perturbed inlier and outlier data and demonstrate improved robustness compared to prior methods. Nevertheless, both methods require adversarial examples for training the respective detector. 

Another line of research has found that as neural networks become more robust, the interpretability of their saliency maps increases \citep{Tsipras19Interpretability,Etmann19Saliency}. \citet{Gu2019Saliency} propose enhanced Guided Backpropagation and show that the classifications of adversarial images can be explained by saliency-based methods. \citet{Ye2020Saliency} demonstrate that the saliency maps of adversarial and benign examples exhibit different properties and utilize this behaviour to detect adversarial attacks. However, \citet{Dombrowski19SaliencyGeometry} observe that explanation-based methods can be manipulated by adversarial attacks as well, which limits the robustness of these methods.

\section{Geometric Gradient Analysis}

In this section we first introduce the necessary mathematical notation and describe the proposed geometric gradient analysis (GGA) method. Then, necessary and sufficient conditions for local minima in the loss function using non-local gradient information are given to further motivate the geometrical gradient analysis.


Let $(x,y)$ be a pair consisting of an input sample $x \in \mathbb{R}^d$ and its corresponding class label $y \in \{1, \dots, C\}$ in a supervised classification task. We denote by $F_{\theta}$ a neural network parametrized by the parameter vector $\theta \in \Theta$, and by $\hat{k}$ the class predicted by the neural network for a given sample $x$. We define $\mathcal{L}(F_{\theta} (x), y)$ as the loss function of the neural network. The GGA method can be summarized as follows. We first define $s_{i}(x) \in \mathbb{R}^d$ as the saliency map of the $i$-th class for a given sample $x$ as
\begin{equation}
    s_{i}(x) = \operatorname{sgn}\left(\nabla_{x} \mathcal{L} \left( F_{\theta} (x), i \right)\right),
\end{equation}
where $\operatorname{sgn}$ indicates the element-wise sign operation. As common for adversarial attacks \citep{Goodfellow2015,Madry2018} we use the sign of the gradient instead of utilizing the gradient directly. This has shown to be effective for approximating the direction which will maximize the loss w.r.t.\ to the respective class \citep{Goodfellow2015,Madry2018} and has been more effective for GGA as well in our experiments. 
Omitting the dependency on $x$, the cosine similarity matrix $\text{CSM} \equiv \text{CSM}(x)$, for a given sample $x$ is defined as 
\begin{equation}
    \text{CSM} = \left( c_{ij} \right) \in \mathbb{R}^{C \times C}, c_{ij} = \frac{s_i \cdot s_j}{|s_i| |s_j|}
\end{equation}
where $i, j \in \{1, ..., C\}$ and $c_{ij}$ represent the cosine similarity between the two saliency maps $s_i$ and $s_j$. In contrast to previous methods, which rely solely on the saliency w.r.t.\ the predicted class, GGA takes into account the geometric properties between the saliency maps of all possible output classes. Considering multiple saliency maps simultaneously makes GGA more difficult to attack. To fool the trained neural network as well as the GGA detector, an attacker must cause a misclassification while simultaneously retaining the geometric properties between the saliency maps of all output classes. As we demonstrate in Section \ref{section:loss_landscape}, after network training correctly classified inputs $x$ are mostly mapped onto a local minimum of the loss landscape w.r.t. the predicted class $\hat{k}$. For these correctly classified samples the saliency maps of non-predicted classes $s_{i}, s_{j}$, $i, j \neq \hat{k}$, point away from the local minimum and exhibit a high average cosine similarity. In contrast, incorrectly classified samples leave the vicinity of these local optima and show different saliency maps for different classes and thus a lower average cosine similarity and more variance between the saliency maps. 

\paragraph{Necessary and sufficient conditions for local minima of the loss function:}

To further motivate the analysis of the geometry of gradients in the input space of neural networks we introduce a property that lets us identify if a given data point lies on a local minimum of the loss landscape. First, we observe that the following holds.

\begin{theorem} \label{thm:zeta}
Let $\zeta_x$ be defined by
\begin{align} \label{eq:quotient}
    \zeta_x\equiv\zeta_x(\tilde x) :=  \frac{\left\langle -\nabla_x \mathcal{L}(F_\theta(\tilde{x}),i), x-\tilde{x} \right\rangle}{|\nabla_x \mathcal{L}(F_\theta(\tilde{x}),i)||x-\tilde{x}|},\quad \tilde x\neq x.
\end{align}
The point $x$ is a local minimum of $\mathcal{L}(F_\theta(\cdot),i)$ if and only if
\begin{align}\label{ineq:quotient}
    0 \leq \liminf_{|x-\tilde x|\to 0}\zeta_x(\tilde x) \leq  \limsup_{|x-\tilde x|\to 0}\zeta_x(\tilde x) \leq 1. 
\end{align}
\end{theorem}

The proof of Theorem~\ref{thm:zeta} is divided in two steps. We first prove that \eqref{ineq:quotient} is necessarily met if the function $x\mapsto \mathcal{L}(F_\theta(x),i)$ attains a local minimum in $x$.
This follows from
\begin{lemma}\label{lem:necessity}
Let $f:\R^d\to\R$ be a $C^1$-function and let $x$ be a local minimum of $f$.
Then it holds
\begin{align*}
    0 &\leq \liminf_{|x-\tilde x|\to 0}\frac{\langle-\nabla f(\tilde x),x-\tilde x\rangle}{|\nabla f(\tilde x)||x-\tilde x|} \\
    &\leq \limsup_{|x-\tilde x|\to 0}\frac{\langle-\nabla f(\tilde x),x-\tilde x\rangle}{|\nabla f(\tilde x)||x-\tilde x|} \leq 1.
\end{align*}
\end{lemma}
\begin{proof}
Taylor expanding around $\tilde x$ gives
\begin{align*}
    f(x) &= f(\tilde x) + \langle \nabla f(\tilde x), x-\tilde x\rangle + o(|x-\tilde x|),
\end{align*}
which can be reordered to
\begin{align*}
    \frac{f(\tilde x) - f(x)}{|x-\tilde x|} = \frac{\langle-\nabla f(\tilde x),x-\tilde x\rangle}{|x-\tilde x|} + o(1).
\end{align*}
If $x$ is a local minimum, one obtains
\begin{align*}
    0 \leq \liminf_{|x-\tilde x|\to 0}\frac{\langle-\nabla f(\tilde x),x-\tilde x\rangle}{|x-\tilde x|}
\end{align*}
which directly implies the desired inequality.
\end{proof}
Non-negativity of the cosine similarity in \eqref{eq:quotient} can also be brought into correspondence with positive semi-definiteness of the Hessian of $f$ which follows from
\begin{lemma}\label{lem:hessian}
Let $f:\R^d\to\R$ be a $C^2$-function.
Then for all vectors $e\in\R^d$ with $|e|=1$ it holds
\begin{align*}
    \lim_{r\to 0} \left\langle\frac{\nabla f(x+re)-\nabla f(x)}{r},e\right\rangle =\left\langle Hf(x)e,e\right\rangle.
\end{align*}
\end{lemma}
\begin{proof}
We compute
\begin{align*}
    &\phantom{=}\left\langle\frac{\nabla f(x+re)-\nabla f(x)}{r},e\right\rangle\\
    &=\left\langle\frac{1}{r}\int_0^r\frac{\d}{\d t}\nabla f(x+te)\d t,e\right\rangle \\
    &=\left\langle\frac{1}{r}\int_0^rHf(x+ty)e\d t,e\right\rangle 
\end{align*}
where $Hf=\left(\partial_i\partial_j f\right)_{i,j}$ denotes the Hessian matrix of $f$.
Since $f$ is a $C^2$-function, the integral $\frac{1}{r}\int_0^rHf(x+ty)\d t$ converges to $Hf(x)$ as $r\to 0$.
Therefore, one obtains
\begin{align*}
    \lim_{r\to 0}\left\langle\frac{1}{r}\int_0^rHf(x+ty)e\d t,e\right\rangle=\left\langle Hf(x)e,e\right\rangle.
\end{align*}
\end{proof}
We can now proceed to the proof of Theorem~\ref{thm:zeta}.
\begin{proof}[Proof of Theorem~\ref{thm:zeta}]
Applying Lemma~\ref{lem:necessity} to $f(x):=\mathcal{L}(F_\theta(x),i)$ shows that \eqref{ineq:quotient} is necessary for $x$ to be a local minimum.

For the converse direction we argue as follows:
First, we note that \eqref{ineq:quotient} implies that $x$ is a critical point with $\nabla f(x)=0$.
Otherwise one could set $\tilde x=x-t\nabla f(x)$ with $t>0$ and obtain
\begin{align*}
    \frac{\langle-\nabla f(\tilde x),x-\tilde x\rangle}{|\nabla f(\tilde x)||x-\tilde x|} 
    &= \frac{\langle-\nabla f(\tilde x),\nabla f(x)\rangle}{|\nabla f(\tilde x)||\nabla f(x)|}\to -1,
\end{align*}
as $|x-\tilde x|\to 0$, since $\nabla f$ is continuous. 
This is a contradiction to \eqref{ineq:quotient} and hence $\nabla f(x)=0$.

This allows us to compute
\begin{align*}
    &\phantom{=}\frac{\langle-\nabla f(\tilde x),x-\tilde x\rangle}{|\nabla f(\tilde x)||x-\tilde x|} \\
    &=\frac{\langle\nabla f(x)-\nabla f(\tilde x),x-\tilde x\rangle}{|\nabla f(x) - \nabla f(\tilde x)||x-\tilde x|} \\
    &=\left\langle\frac{\nabla f(x)-\nabla f(\tilde x)}{|x-\tilde x|},\frac{x-\tilde x}{|x-\tilde x|}\right\rangle\frac{|x-\tilde x|}{|\nabla f(x)-\nabla f(\tilde x)|}.
\end{align*}
Hence, if this expression is asymptotically non-negative for all $\tilde x$ converging to $x$, we can choose arbitrary $e\in\R^d$ with $|e|=1$, define $\tilde x=x+re$ and apply Lemma~\ref{lem:hessian} to get
\begin{align*}
    0 \leq \lim_{r\to 0}\left\langle \frac{\nabla f(x+re)-\nabla f(x)}{r},e \right\rangle = \langle Hf(x)e,e\rangle.
\end{align*}
Since $e$ was arbitrary, this means that $x$ is a local minimum of the loss $f$.
\end{proof}

\section{Experiments} \label{sec:Experiments}

In this section we first analyze the characteristics of the loss landscape of neural networks in case of correct and incorrect classifications. Further, we demonstrate the effectiveness of GGA on several benchmark data sets. We compare GGA to two other methods which also not necessarily require any re-training of the neural network and do not utilize adversarial examples for training the outlier/adversarial detector. Namely, the method proposed in \citep{Hendrycks17} (called \emph{Baseline} in the following) and the \emph{ODIN method} \citep{Liang18}.  We additionally consider the method proposed by \citet{Lee18} (called \emph{Maha} in the following), which requires the detector to be trained with adversarial examples. 



\subsection{Evaluation Metrics}

We use common evaluation metrics for the assessment of the OOD detection methods \citep{Hendrycks17,Liang18}. This includes: \textbf{1) TNR ($95\%$ TPR)} true negative rate at $95\%$ true positive rate, \textbf{2) AUPR}: the area under precision recall curve, which we report both for the in-distribution and OOD as positives as AUPR-In and AUPR-Out, respectively, and \textbf{3) AUROC}: the area under the receiver operating characteristic curve.

\subsection{Setup} 

In the following we give an overview of general hyperparameters used for the performed experiments. This includes a description of the data sets, neural network models, and the features we extract from the CSMs. Finally, we describe the threat model of the adversarial attacks.

\subsubsection{Data and Architectures}

 We split each data set into predefined training and testing sets. Additionally, we used $10\%$ of the training data as the validation set for self-trained models. All self-trained models were trained by minimizing the cross-entropy loss using SGD with Nesterov momentum ($0.9$) and a batch size of $128$. We used a step-wise learning rate schedule that divides the learning rate by five at 30\%, 60\%, and 80\% of the total training epochs. The following classification data sets were used to evaluate the proposed method.

\textbf{MNIST} \citep{LeCun98} consists of greyscale images of handwritten digits each of size $28\times28\times1$ ($60,000$ training and $10,000$ test) and is a common benchmark for outlier detection and adversarial robustness. We trained a basic CNN architecture as in prior work \citep{Madry2018}. This architecture consists of four convolutional layers with $32$, $64$, and $128$ filters and two fully-connected layers with $100$ and $10$ output units, respectively. We used ReLU for the activation functions between each layer. We used a learning rate of $0.1$ and trained for $10$ epochs, where the validation accuracy converged.  

\textbf{CIFAR10} \citep{Krizhevsky2009} consists of RGB color images, each of size $32\times32\times3$, with $10$ different labels ($50,000$ training and $10,000$ test). For CIFAR10 we used a ResNet56 \citep{He2016}. All images from the CIFAR10 data set were standardized and random cropping and horizontal flipping were used for data augmentation during training as in \citep{He2016}. 

\textbf{CIFAR100} \citep{Krizhevsky2009} has the same properties as CIFAR10 but is considerably more difficult as it contains $100$ instead of only $10$ classes. For CIFAR100 we used a pre-trained PreResNet164 \citep{pytorch_models} and otherwise the same configurations as for CIFAR10.

\textbf{UCR ECG (ID 49)} \citep{Dau2018UCR} is a time series classification data set with $42$ different classes ($1800$ training and $1965$ test). It contains non-invasive electrocardiogram (ECG) recordings of fetuses with a length of $750$ time steps each. We consider this data set in addition to the computer vision data sets for a basic benchmark of the proposed GGA method on time series classification tasks. We trained a basic CNN architecture consisting of three convolutional layers with $128$, $256$, and $128$ filters and one fully-connected layer with $42$ output units. We used batch normalization and ReLU as activation function between each layer. We used a learning rate of $0.01$ and trained for $100$ epochs.

\subsubsection{Out-Of-Distribution Data Sets}

We consider the respective test set of the training data as in-distribution data and suitable realistic images from other data sets as OOD. Additionally, for every data set we create two synthetic noise data sets as OOD data as done in prior work \citep{Hendrycks17,Liang18}. In the following list the respective in-distribution data sets are put in brackets after the OOD data sets:

\textbf{Fashion-MNIST (OOD for MNIST)}: \citep{Xiao2017} consists of greyscale images of $10$ different types of clothing, each of size $28\times28\times1$ ($10,000$ test).

\textbf{SVHN (OOD for CIFAR10, CIFAR100)}: \citep{Netzer2011} consists of color images of $10$ different types of street view digit, each of size $32\times32\times3$ ($26,032$ test).

\textbf{Uniform Noise (OOD for all data sets)}: the synthetic uniform noise data set consists of $10,000$ images, where each pixel value is drawn i.i.d. from a uniform distribution in $[0, 1]$.

\textbf{Gaussian Noise (OOD for all data sets)}: the synthetic Gaussian noise data set consists of $10,000$ images, where each pixel value is drawn i.i.d. from a uniform distribution with unit variance.

\subsubsection{Geometric Gradient Analysis Features and Prediction} \label{sec:GGAFeatures}

To identify untrustworthy predictions with the GGA method, we first generate the respective CSM for a given sample $x$. Then, we compute simple features from the CSMs and use them for training a simple outlier detector. Let $\hat{k}$ be the index associated with the class predicted by the neural network $F_\theta$ for a given sample $x$. By exploiting the symmetry of the cosine similarity matrix CSM, and observing that the elements of the main diagonal are all equal to $1$, we can restrict the analysis to the set $S$ to the elements above the main diagonal, i.e., $S = \{c_{ij} \}_{i < j}$.
We compute five basic statistical features (\textit{mean, maximum, minimum, standard deviation, and energy}) separately for two different sets $S_1 = S \cap \{c_{ij} \}_{i, j \ne k}$ and $S_2 = S \cap  \{c_{ij} \}_{i = \hat{k} \vee j = k }$. These statistics constitute the ten features $f_{1-10}$ provided to the outlier detection model.

\begin{figure}[ht]
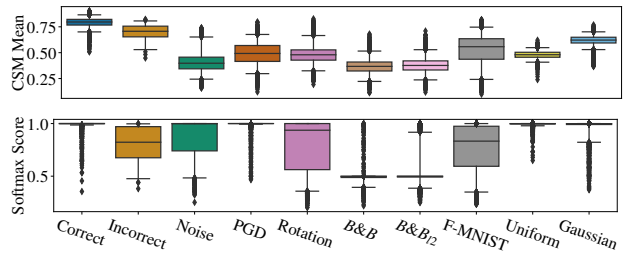

 \begin{subfigure}{0.48\textwidth}
    \resizebox{\textwidth}{!}{\input{Images/BoxPlot_CSM_Mean_no_label.pgf}}
  \end{subfigure}
  \begin{subfigure}{0.48\textwidth}
    \resizebox{\textwidth}{!}{\input{Images/BoxPlot_Softmax_Score.pgf}}
  \end{subfigure}
 \caption{Box-plots of the mean value of the cosine similarity matrices and the softmax scores of the neural network for different data. The boxes show the quartiles of the data set while the whiskers extend to $95\%$ of the distributions. The values are calculated on the \textbf{MNIST} validation set using the \textbf{MNIST} model.}
\label{fig:boxplot}
\end{figure}

Figure \ref{fig:boxplot} exemplifies how the mean value of $S_1$ of a CSM can be used to differentiate between several data classes on the MNIST data set that are not discriminated by the softmax score alone. We exclude the softmax score from the GGA features for better comparison between the methods. For practical applications the softmax score can be used as an additional feature. For all the remaining detection tasks we train a lightweight on-line detector of anomalies (LODA) \citep{Pevny16} with the GGA features of the correctly classified samples of the training set. We chose LODA as it is designed to handle a large number of data points and does not add a noticeable computational overhead to the classification pipeline. For LODA we set the number of random cuts to $100$ for all experiments. The number of random bins was set to $500$ for the CIFAR data sets and $100$ for MNIST and UCR ECG after an evaluation on the validation set. 

\subsubsection{Threat Model} 
Let $\delta \in \mathbb{R}^d$ be an adversarial perturbation. We use a variety of adversarial attacks with different attributes to generate untrustworthy predictions. We only consider successful adversarial attacks that change the classification result as untrustworthy and discard unsuccessful attacks. We employ attacks with different norm constraints ($\ell_2$, $\ell_{\infty}$) such that the adversarial perturbation is smaller than some predefined perturbation budget $||\delta||_p \leq \epsilon$. We set the perturbation budget $\epsilon$ in the $\ell_{\infty}$-norm to $0.3, 8/255, 8/255, 0.1$ for MNIST, CIFAR10, CIFAR100, and UCR ECG, respectively, as in prior work \citep{Madry2018,Fawaz19timeSeries}. For attacks in the $\ell_2$-norm we multiply the allowed perturbation strength by $10$. Furthermore, we use attacks that produce high and low confidence predictions as low confidence adversarial examples have shown to be effective against several detectors \citep{Chen2020robust}. To create high confidence misclassifications we use Projected Gradient Descent (PGD) \citep{Madry2018}. PGD is an iterative attack that tries to maximize the loss w.r.t.\ the original class and subsequently creates perturbations which lead to wrong predictions with high certainty. For the PGD attack we used a step size of $\alpha = \frac{\epsilon}{4}$ and $70$ attack iterations which lead to a success rate of $100\%$ for all models. To create low confidence predictions we employ the B\&B attack \citep{Brendel2019} which creates perturbations at the decision boundary of the attack. For B\&B we used a learning rate of $0.001$ and $100$ iterations. Finally, we consider random rotations between $-45$ and $45$ degrees and uniform noise attacks in the $\epsilon$-ball for non-gradient-based attacks (rotations are naturally omitted for the time-series classification task). 

We create several adaptive attacks that are designed to fool the proposed GGA method \citep{Grosse17faileddetector,Carlini17detect}. With these attacks we aim to obtain cosine similarity matrices which resemble those of correctly classified samples. This means that the cosine similarity between non-predicted classes in $S_1$ should be high on average while the cosine similarities in $S_2$ are small.

\textbf{Targeted attacks}: We use a targeted PGD-based attack to maximize the loss w.r.t.\ to a random target class which is not the ground truth. We argue that such attacks could result in similar saliency maps for all other classes since the attack will optimize the input towards a local minimum of the loss landscape w.r.t.\ to the target class. We employ this attack both with the mean squared error (T-MSE) and the categorical cross-entropy loss (T-SCE). We used the same step size of $\alpha = \frac{\epsilon}{4}$ and $100$ attack iterations for all targeted attacks.

\textbf{Cosine similarity attack (CSA)}: We use a PGD-based attack and additionally add a cosine similarity objective to optimize the perturbation such that the saliency maps of all non-predicted classes align. The loss of the cosine similarity objective is given by:

\begin{equation} \label{eq:csm_attack}
\mathcal{L}^{CSA}(x, \hat{k}) = \frac{1}{|S_1|} \sum_{c_{ij} \in S1} c_{ij}(x).
\end{equation}

For this attack we exchange the ReLU activation functions \citep{Agarap2018ReLU} with Softplus activations \citep{Dugas00Softplus}. This was shown to be an effective way to calculate a second-order gradient to attack the saliency maps of neural networks \citep{Dombrowski19SaliencyGeometry}. We used the same step size as for the other PGD attacks and attack iterations as for the targeted attacks. We achieved the highest success rate of the attack by weighting the CS objective in \eqref{eq:csm_attack} by $0.8$ and the cross-entropy objective with $0.2$. As our cosine similarity attack needs the second-order gradient for optimization it may be prone to common pitfalls, such as gradient obfuscation \citep{Athalye18}. We encourage other researchers to create adaptive attacks that circumvent our method.

\section{Results and Discussion}

\begin{table*}[t]
    \caption{True negative rate in [\%] for different augmentations, OOD data, and attacks. All values are given for a true positive rate of $95\%$. Additionally the AUROC, AUPR-In, and AUPR-Out for all data types is shown.}
    \small
    \centering
    \begin{tabular}{llrrrrrrrrrr}
        \toprule
        Data set & Method & Noise & PGD & Rotation & B\&B & B\&B\textsubscript{L2} & OOD & AUROC & AUPR-In & AUPR-Out \\
        \midrule
         \textbf{MNIST} & \textbf{Baseline} & 45.8 & 3.3 & 59.7 & 99.9 & 99.3 & 55.0 & 86.1 & 46.7 & 97.8 \\
         & \textbf{ODIN} & 97.2 & 2.5 & 92.0 & 93.6 & 89.1 & 69.4 & 88.2 & 39.9 & 98.1  \\
          & \textbf{Maha} & \textbf{100.0} & 12.4 & 93.5 & 98.4 & 99.0 & 97.9 & 92.4 & 88.3 & \textbf{99.9} \\
          & \textbf{Ours} & \textbf{100.0 }& \textbf{98.9} & \textbf{98.2} & \textbf{100.0} & \textbf{100.0} & \textbf{98.1} & \textbf{99.5} & \textbf{97.7} & \textbf{99.9}   \\
         \midrule
         \textbf{CIFAR10} & \textbf{Baseline} & 10.5 & 0.0 & 55.1 & 97.5 & 95.4 & 77.2 & 82.7 & 30.2 & 97.4  \\
          & \textbf{ODIN} & 25.3 & 0.0 & 45.1 & 14.0 & 14.5 & 83.6 & 81.2 & 29.7 & 96.8  \\
          & \textbf{Maha} & 93.1 & 85.9 & 73.2 & 90.8 & 91.3 & \textbf{88.2} & 90.0 & 72.1 & 99.0 \\
          & \textbf{Ours} &\textbf{95.6} & \textbf{92.6} & \textbf{84.5} & \textbf{93.2} & \textbf{93.3} & 84.2 & \textbf{96.3} & \textbf{83.7} & \textbf{99.4}  \\
         \midrule
         \textbf{CIFAR100} & \textbf{Baseline} & 32.4 & 0.0 & 40.1 & 80.7 & 81.5 & 6.7 & 55.2 & 11.3 & 93.6 \\
         & \textbf{ODIN} & 16.8 & 0.0 & 15.6 & 8.5 & 7.9 & 23.3 & 60.2 & 16.3 & 94.1 \\
         & \textbf{Maha} & 93.7 & 81.9 & 77.3 & 52.1 & 55.3 & \textbf{86.2} & 68.2 & 47.1 & 98.4 \\
          & \textbf{Ours} & \textbf{95.1} & \textbf{98.5} & \textbf{95.1} & \textbf{98.1} & \textbf{97.9} & 83.5 & \textbf{98.0} & \textbf{87.5} & \textbf{99.7} \\
         \midrule
         \textbf{UCR ECG} & \textbf{Baseline} & 6.7 & 0.5 & N/A & 1.5 & 1.8 & 0.0 & 11.8 & 6.9 & 74.7 \\
         & \textbf{ODIN} & 0.0 & 0.0 & N/A & 0.0 & 0.0 & 0.0 & 0.0 & 6.4 & 70.7 \\
         & \textbf{Ours} & \textbf{81.5} & \textbf{96.7} & N/A & \textbf{75.9} & \textbf{75.8} & \textbf{100.0} & \textbf{96.9} & \textbf{88.8} & \textbf{99.1} \\
         \bottomrule
    \end{tabular}
    \label{tab:adversarial}
\end{table*}

In the following we summarize and analyze the findings of the experiments used to evaluate the proposed GGA method.  

\subsection{Loss Landscape Analysis} \label{section:loss_landscape}
\begin{figure}[t]
\centering
 \begin{subfigure}{0.4\textwidth}
    \resizebox{\textwidth}{!}{\input{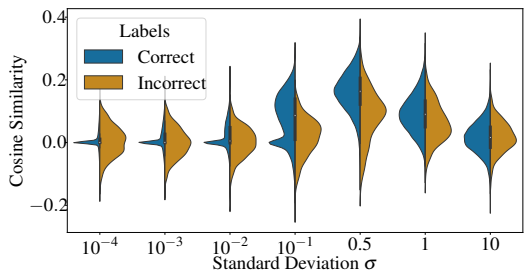}}
  \end{subfigure}
 \caption{Violin plots of  $\zeta_x$ estimates. Correct classifications correspond to a local minimum of the loss. The values are calculated for the validation set of the \textbf{MNIST} data set.}
\label{fig:CS_Statistics}
\end{figure}

In a preliminary experiment we investigated if correctly classified samples $x$ with predicted label $i$ lie on local minima of the loss landscape while incorrectly classified samples do not. Therefore, we empirically investigate the properties of $\zeta_x$ defined in \eqref{eq:quotient}.
Since $\zeta_x$ is a cosine similarity, it holds that $\zeta_x\in [-1,1]$. To test our hypothesis we estimate $\zeta_x$ in a neighborhood of a sample $x$ and check whether it is non-negative. To generate points close to $x$ we add i.i.d.\ Gaussian noise with standard deviation $\sigma>0$. Following this procedure, we calculated the statistics of $\zeta_x$ in \eqref{eq:quotient} with $1,000$ injections per sample on the MNIST validation set. The results are displayed in Figure~\ref{fig:CS_Statistics}. In the direct vicinity of the original sample the gradients are mostly orthogonal to noise for correct classifications, indicating that the corresponding samples lie in a relatively wide minimum of the loss. Incorrect classifications show a significantly larger spread of the values of $\zeta_x$, which corresponds to a saddle point. For increasing standard deviation $\zeta_x$ is larger for correct classifications than for incorrect ones, supporting the hypothesis that the former correspond to a local minimum. When the samples $\tilde x$ are too far from $x$ the value $\zeta_x$ becomes normally distributed around zero. 

While the quantity in \eqref{eq:quotient} already serves as a good indicator of misclassifications adversarial attacks could move a sample towards a local minimum of the loss landscape with respect to the predicted class. Hence, we additionally consider the gradient directions w.r.t.\ other classes with the GGA method, which makes the detection of misclassifications induced by adversarial attacks substantially more robust. 

\subsection{Out-Of-Distribution Detection and Adversarial Attacks} \label{sec:ood_adv}

First, we studied the detection performance on OOD data and adversarial attacks as described in the previous section. The results are summarized in Table \ref{tab:adversarial}. As reported in prior work the baseline and ODIN method fail to identify adversarial attacks. In contrast, the proposed GGA shows high identification performance for all attacks. The augmentations which were most difficult to detect for the computer vision tasks were rotations. UCR ECG noise and the B\&B attacks were most difficult to detect. For the detection of OOD data the GGA method achieves worse results than the Mahalinobis distance based approach (Maha) in some cases \citep{Lee18}. However, in contrast to GGA, Maha requires additional finetuning of the OOD and adversarial detector on OOD data and adversarial examples, respectively.

\subsection{Adaptive Adversarial Attacks}

Next, we studied the detection performance on adaptive adversarial attacks which were specifically designed to fool the GGA detector. As seen in Table \ref{tab:adaptive_robust}, the proposed GGA shows high identification performance on all adaptive attacks. The targeted PGD attacks (T-SCE, T-MSE) show a higher success rate than the untargeted PGD attack. Using the softmax cross-entropy loss for the targeted attack was more effective in our experiments. We observed that we can successfully increase the cosine similarity between non-predicted classes in the cosine similarity matrices with the cosine similarity attack (CSA). CSA achieves considerably higher success-rate than the standard PGD attack. However, combining this objective with the goal to induce misclassifications seems to be ineffective. A higher weight for the cosine similarity objective results in considerably fewer misclassifications and vice versa. The CSA attack was only able to induce a misclassification on $561$, $1354$, and $857$ out of $10,000$ samples for the MNIST, CIFAR10, and CIFAR100 data sets, respectively. In contrast, the untargeted and targeted PGD attacks achieved $100\%$ success rate and led to a misclassification on $10,000$ out of $10,000$ images on all data sets.

\begin{table}[t]
    \caption{Identification accuracy [\%] for different adaptive attacks for the proposed GGA. All values are given for a TPR of $95\%$}
    \centering
    \begin{tabular}{lrrr}
        \toprule
        Data set & T-SCE & T-MSE & CSA  \\
        \midrule
         \textbf{MNIST} & 95.4 & 97.3 & 72.1 \\
         \midrule
         \textbf{CIFAR10} & 90.3 & 91.5 & 69.7 \\
         \midrule
         \textbf{CIFAR100} & 96.9 & 97.8 & 71.6 \\
         \bottomrule
         \end{tabular}
    \label{tab:adaptive_robust}
\end{table}

\subsection{Gradient Obfuscation}

\begin{figure}[ht]
 \begin{subfigure}{0.4\textwidth}
    \resizebox{\textwidth}{!}{\input{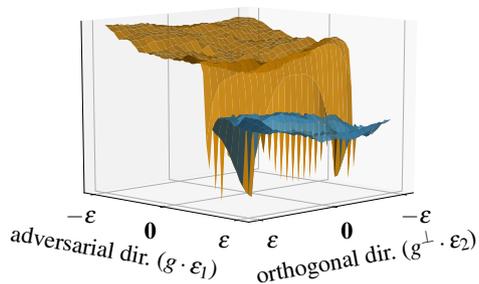}}
  \end{subfigure}
 \caption{Landscape of the mean value of cosine similarity maps centered around a clean sample $x$. We calculate the loss value for sample $x + \epsilon_{1} \cdot \gamma + \epsilon_{2} \cdot \gamma^\perp$ where $\gamma$ is the direction of a successful adversarial attack and $\gamma^\perp$ a random orthogonal direction. The different colors indicate the predicted class of the neural network.}
\label{fig:csm_loss_landscape}
\end{figure}

Prior work demonstrates that defense mechanisms that are apparently robust to adaptive attacks can often be circumvented with another or simpler optimization objectives \citep{Athalye18,Tramer2020}. Complex objectives often result in noisy loss landscapes with unreliable gradient information. To evaluate if this phenomenon applies to the CSA attack, we further inspect the behaviour of the CSM features over a wide variety of perturbed data points. In particular, we inspect the behaviour of the objective in \eqref{eq:csm_attack} along the direction of a successful adversarial perturbation ($g$) and a random orthogonal direction ($g^\perp$) originating from a clean sample. This results in a three-dimensional map where the $x$-axis $g$ and $y$-axis $g^\perp$ describe the perturbation of the current data point, while the $z$-axis shows the value described in \eqref{eq:csm_attack}. A representative map for an individual sample of the MNIST data set is shown in Figure \ref{fig:csm_loss_landscape}. The predicted label of the classifier is color coded, where the upper plateau in orange corresponds to the ground truth class while the lower plateau in blue corresponds to the class predicted after the adversarial attack. Near the decision boundary the CSM characteristics fluctuate as the gradient directions between saliency maps of different classes start to diverge. It can be seen that the mean value of the CSMs is a stable indicator of the classifier decision. Furthermore, the smoothness of the map indicates that the mean value can be utilized as an objective for an adaptive attack. 

\subsection{Enhancing the Efficiency}

The main computational cost of GGA is given by the preliminary computation of the saliency maps for each reference class. Here, we show that it is possible to rely on a partial computation of the CSM with only the top-$N$ predicted classes. This allows GGA to scale to data sets with a large number of output classes. Table \ref{tab:top_n} demonstrates the detection performance for the same adversarial attacks and outlier data as used in Section \ref{sec:ood_adv} for CSMs which are computed with the top-$N$ predictions only. Even for the case of only $5\%$ of the original saliency maps used to calculate the CSMs the performance degrades only marginally for all detection tasks. We observed that the cosine similarity between the gradients of the predicted class and the non-predicted classes is mostly sufficient for the detection of untrustworthy predictions. We argue that this enables the algorithm to perform well with largely reduced CSMs. In our experiments the partial CSMs performed similarly on the adaptive attacks as well. Note that the computation of the CSMs can be parallelized, which results in no time overhead between partial and full CSMs when sufficient memory is used for the calculation. In practice, the computational overhead can be adjusted based on the required detection performance and available resources.

\begin{table}[t]
    \caption{Detection performance for cosine similarity maps which are calculated with only the top-$N$ prediction of the classifier. AUROC, AUPR-In, and AUPR-Out in [\%] for different augmentations, OOD data, and attacks are shown.}
    \centering
    \begin{tabular}{lrrr}
        \toprule
        CIFAR100 & AUROC & AUPR-In & AUPR-Out  \\
        \midrule
        \textbf{top-100 (all)} & 98.0 & 87.5 & 99.7 \\
        \midrule
        \textbf{top-10} & 98.0 & 86.9 & 99.7 \\
         \midrule
        \textbf{top-5} & 97.7 & 85.6 & 99.7 \\
        \bottomrule
        \end{tabular}
    \label{tab:top_n}
\end{table}

\section{Conclusion}

In this paper we proposed a novel geometric gradient analysis (GGA) method, which is designed to identify out-of-distribution data and adversarial attacks in neural networks. The proposed method does not require re-training of the neural network model and can be used with any pre-trained model. We first mathematically motivated the proposed GGA by relating geometric properties in the input space of a neural network to characteristics of the loss landscape. Next, we demonstrated that GGA achieves competitive performance for outlier detection. Furthermore, we showed that GGA effectively detects state-of-the-art and adaptive adversarial attacks. Finally, we demonstrated how GGA can be efficiently implemented for data sets with a large number of output classes. Future work will explore the end-to-end training of a GGA-based detector with the CSMs. 






\bibliography{uai2021-template}

\end{document}